\documentclass[letterpaper, 10 pt, conference]{ieeeconf}  

\IEEEoverridecommandlockouts                              

\makeatletter
\let\NAT@parse\undefined
\makeatother

\usepackage[numbers]{natbib}
\usepackage[utf8]{inputenc} 
\usepackage[T1]{fontenc}    
\usepackage{hyperref}       
\usepackage{url}            
\usepackage{booktabs}       
\usepackage{amsmath}
\usepackage{amsfonts}       
\usepackage{nicefrac}       
\usepackage{microtype}      
\usepackage{xcolor}         

\colorlet{mygray}{gray!50!darkgray}

\usepackage{bbm}
\usepackage{graphicx}
\usepackage{wrapfig}
\usepackage{cite}
\usepackage{makecell}
\usepackage{comment}
\usepackage{cleveref}
\usepackage{soul}
\usepackage[labelfont=bf,font=small]{caption}
\usepackage{subcaption}
\usepackage{thm-restate}

\usepackage{siunitx}
\usepackage{titlesec}

\usepackage[export]{adjustbox}
\usepackage{glossaries}
\glsdisablehyper
\setacronymstyle{long-sc-short}

\usepackage{tikz}

\DeclareMathOperator*{\sign}{sign}
\DeclareMathOperator*{\JSD}{JSD}

\DeclareMathOperator*{\minADE}{minADE}
\DeclareMathOperator\supp{supp}
\DeclareMathOperator\KL{KL}
\newcommand{\bb}[1]{\mathbb{#1}}
\newcommand{\bs}[1]{\boldsymbol{#1}} 
\newcommand\bbE{\ensuremath{\mathbb{E}}} 
\newcommand\pol{\ensuremath{\bs{\pi}_{\bs{\theta}}}} 
\newcommand\e{\ensuremath{\bs{e}_{\bs{\theta}}}} 
\newcommand\p{\ensuremath{p_{\bs{\theta}}}} 
\newcommand\pt{\ensuremath{p_{\bs{\bar\theta}}}} 
\newcommand\g{\ensuremath{\bs{g}}} 
\newcommand\gh{\ensuremath{\bs{\hat{g}}}} 
\newcommand\Gh{\ensuremath{{\hat{G}}}} 
\renewcommand\a{\ensuremath{\bs{a}}} 
\newcommand\ah{\ensuremath{\bs{\hat{a}}}} 
\newcommand\A{\ensuremath{{A}}} 
\newcommand\Ah{\ensuremath{{\hat{A}}}} 
\newcommand\s{\ensuremath{\bs{s}}} 
\newcommand\sh{\ensuremath{\bs{\hat{s}}}} 
\renewcommand\S{\ensuremath{{S}}} 
\newcommand\x{\ensuremath{\bs{\xi}}} 
\newcommand\X{\ensuremath{{\Xi}}} 
\newcommand\tauh{\ensuremath{\hat{\tau}}} 
\newcommand\Das{\ensuremath{D_{\bs{\phi}}(\a_t, \s_t)}} 
\newcommand\Dash{\ensuremath{D_{\bs{\phi}}(\ah_t, \sh_t)}} 

\newcommand\name{\texttt{RTC}}
\newcommand\nameC{\texttt{RTC-C}}
\newcommand\nameD{\texttt{RTC-D}}
\newcommand\HNoPT{\texttt{NaiveHierarchy}}

\newcommand\MGAIL{\texttt{MGAIL}}
\newcommand\Symphony{\texttt{Symphony}}
\newcommand\InfoMGAIL{\texttt{InfoMGAIL}}

\newcommand\BC{\texttt{BC}}
\newcommand\fullname{Robust Type Conditioning} 
\newcommand\envname{Double Goal Problem} 

\newcommand\Ladv{\ensuremath{\mathcal{L}_{\text{adv}}}} 
\newcommand\Lrec{\ensuremath{\mathcal{L}_{\text{rec}}}} 
\newcommand\Lvae{\ensuremath{\mathcal{L}_{\text{vae}}}} 
\newcommand\Lkl{\ensuremath{\mathcal{L}_{\text{kl}}}} 
\newcommand\LHuber{\ensuremath{\mathcal{L}_{\text{Huber}}}} 

\newcommand\mc[2]{\ensuremath{\mathcal{#1}_{\text{#2}}}} 

\newacronym{ldf}{LfD}{learning from demonstration}
\newacronym{il}{IL}{imitation learning}
\newacronym{rtc}{RTC}{\emph{Robust Type Conditioning}}

\title{\LARGE \bf
Hierarchical Imitation Learning for Stochastic Environments
}

\author{Maximilian Igl$^{*}$, Punit Shah$^{*}$, Paul Mougin$^{*}$, Sirish Srinivasan$^{*}$, Tarun Gupta$^{\dagger}$, Brandyn White$^{*}$,\\ Kyriacos Shiarlis$^{*}$, Shimon Whiteson$^{*}$
\thanks{* Waymo Research, $\dagger$ U. of Oxford. Work performed during internship.}
}

\begin{document}

\maketitle
\thispagestyle{empty}
\pagestyle{empty}

\begin{abstract}
Many applications of imitation learning require the agent to generate the full distribution of behaviour observed in the training data.
For example, to evaluate the safety of autonomous vehicles in simulation, accurate and diverse behaviour models of other road users are paramount.
Existing methods that improve this \emph{distributional realism} typically rely on hierarchical policies.
These condition the policy on \emph{types} such as goals or personas that give rise to multi-modal behaviour.
However, such methods are often inappropriate for stochastic environments where the agent must also react to external factors:
because agent types are inferred from the observed future trajectory during training, these environments require that the contributions of internal and external factors to the agent behaviour are disentangled and only internal factors, i.e., those under the agent's control, are encoded in the type.
Encoding future information about external factors leads to inappropriate agent reactions during testing, when the future is unknown and types must be drawn independently from the actual future.
We formalize this challenge as distribution shift in the conditional distribution of agent types under environmental stochasticity.
We propose \emph{\fullname{}} (\name), which eliminates this shift with adversarial training under randomly sampled types.
Experiments on two domains, including the large-scale \emph{Waymo Open Motion Dataset}, show improved distributional realism while maintaining or improving task performance compared to state-of-the-art baselines. 
\end{abstract}

\section{Introduction}

Learning to imitate behaviour is crucial when reward design is infeasible \citep{amodei2016concrete,hadfield2017inverse,fu2018learning,everitt2021reward}, for overcoming hard exploration problems \citep{rajeswaran2017learning,zhu2018reinforcement}, and 
for realistic modelling of dynamical systems with multiple interacting agents \citep{farmer2009economy}.
Such systems, including games, driving simulations, and agent-based economic models, often have known state transition functions, but require accurate agent models to be realistic. 
For example, for driving simulations, which are crucial for accelerating the development of autonomous vehicles \citep{suo2021trafficsim,igl2022symphony}, faithful reactions of all road users are paramount.
Furthermore, it is not enough to mimic a single mode in the data; instead, agents must reproduce the full distribution of behaviours
to avoid sim2real gaps in modelled systems \citep{grover2018learning,liang2020agent}.

\begin{figure}[t]
    \centering
    \begin{subfigure}{0.99\linewidth}
        \includegraphics[width=\textwidth]{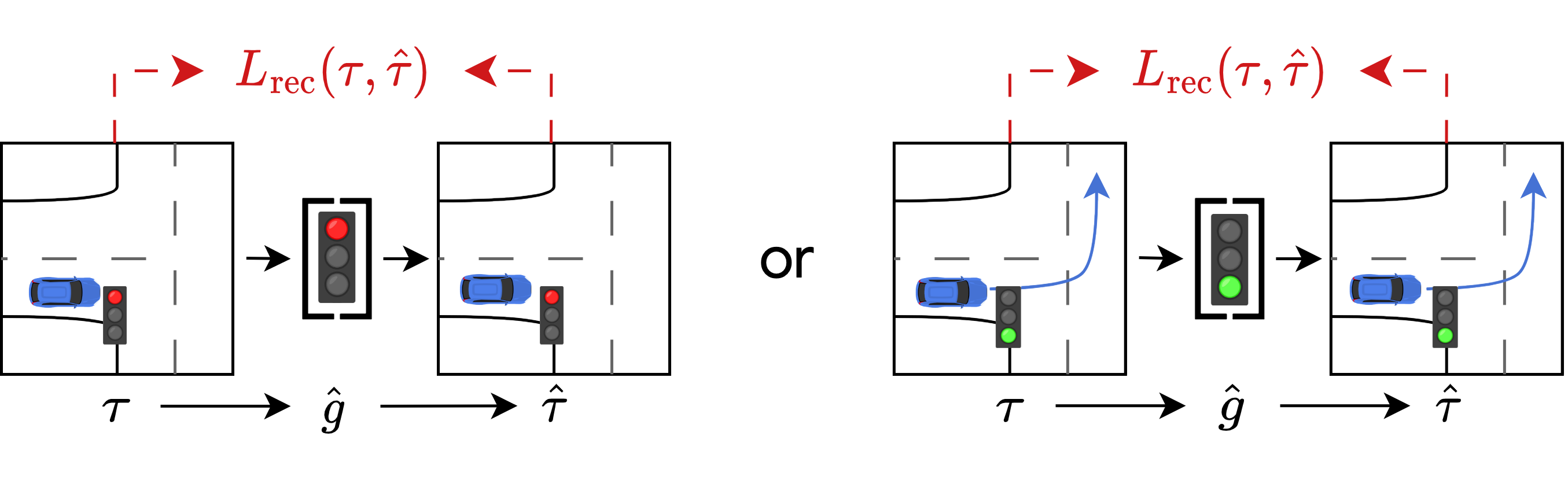}
        \vspace{-1.5em}
        \caption{Training}
        \centering
    \end{subfigure}
    \begin{subfigure}{0.99\linewidth}
        \vspace{1em}
        \includegraphics[width=\textwidth]{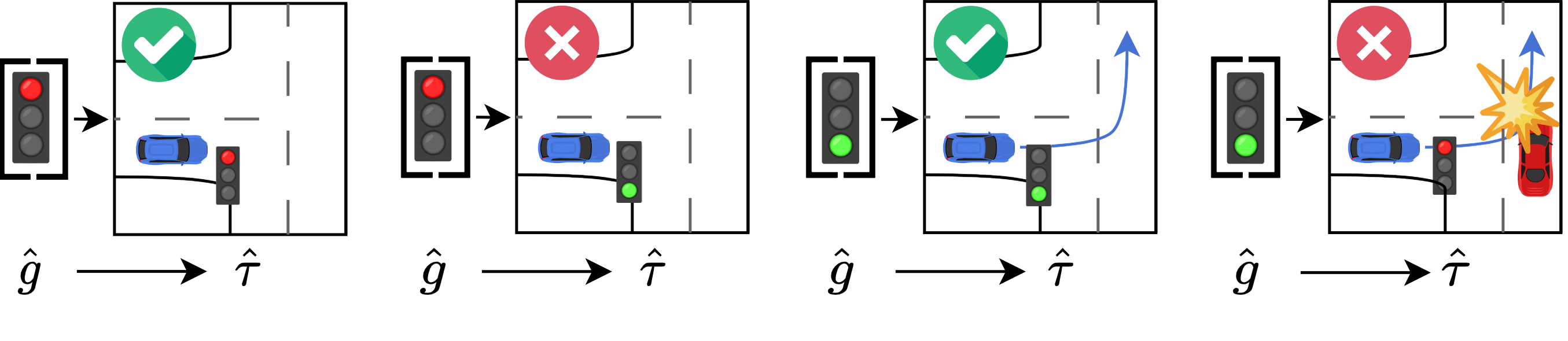}
        \vspace{-1.5em}
        \caption{Testing}
        \centering
    \end{subfigure}
    \caption{
    Example highlighting how stochastic environments can cause out-of-distribution issues for hierarchical policies.
    \textbf{\emph{Top}}: During training, the latent $\gh$ is inferred from the future trajectory $\tau$ in the data using an encoder $\e(\gh|\tau)$.
    In this example, it captures the driving direction and whether the light turns green.
    The policy $\pol(\ah|\gh, \s)$ `decodes' $\gh$ by acting in the environment to generate $\tauh$.
    The reconstruction loss $\Lrec$ penalises differences between $\tau$ and $\tauh$, training the policy to follow $\gh$.
    \textbf{\emph{Bottom}}: During testing, without access to the future, the latent $\gh$ must be sampled randomly from a prior $\p(\gh)$ which was trained to match the \emph{marginal} distribution of possible latents, i.e., it randomly samples red or green lights and possible driving directions. 
    This can cause issues such as collisions when the random latent and the environment do not match. 
    Because the prior cannot know the future, it might sample a red light while the real traffic light turns green (2nd example) or, worse, it might wrongly sample a green light, possibly leading to collisions (last example) if the agent follows the latent $\gh$ as it was trained to do.
    On the other hand, random sampling of agent-\emph{internal} decisions such as driving directions is unproblematic as these do not make assumptions about the future environment.
    }
    \vspace{-1em}
    \label{fig:first_page}
\end{figure}

Current \gls{il} methods fall short of achieving such \emph{distributional realism}: while they are capable of generating individual trajectories that are realistic, they fail to match the full distribution of observed behaviour.
Indeed, the adversarial training objective which enables state-of-the-art performance of most current \gls{il} methods is known to be prone to mode dropping in practice \citep{wang2017robust,lucic2018gans,creswell2018generative}, even though it optimises a distribution-matching objective in principle \citep{goodfellow2020generative,ho2016generative,baram2016model}. 
Furthermore, progress on distributional realism is hindered by a lack of suitable benchmarks, with most relying on unimodal data and only evaluating task performance as measured by rewards, but not mode coverage or recall.
By contrast, many applications, such as agent modeling for autonomous vehicles, require distributional realism in addition to good task performance. 
Consequently, our goal is to improve distributional realism while maintaining strong task performance.
\looseness=-1

To mitigate mode collapse and improve distributional realism in complex environments, previous work uses hierarchical policies in an autoencoder-like framework \citep{wang2017robust,suo2021trafficsim,igl2022symphony,xu2022bits}.
During training, an encoder infers goals from observed future trajectories and the agent, conditioned on those goals, strives to imitate the original trajectory. 
At test time, a prior distribution proposes distributionally realistic goals, without requiring access to privileged future information.
We refer to these goals as an agent's inferred \emph{type} since it can express not only goals, but many agent characteristics responsible for multi-modal behaviour, such as persona, goal, or strategy.

However, as we show in \cref{sec:problem}, using such hierarchical policies in stochastic environments can create a distribution shift between training and testing, possibly leading to out-of-distribution inputs and reduced performance.
Unfortunately, the autoencoder training does not prevent extrinsic information from being encoded.
Intuitively, the type should only capture agent-intrinsic choices that are under the agent's control. 

Consider a car waiting at an intersection (see \cref{fig:first_page}).
During training, because the agent's type (e.g., goal and driving style) are not directly observed, they must be inferred from its future trajectory.
However, this inferred type might not only capture their goal and driving style, but also external factors out of the agent's control, such as the time until the traffic light turns green.
Even innocuous seeming type representations can leak external information; for example, a goal location extracted from the future trajectory can leak information about waiting times based on its distance to the starting position.

Capturing information about external events in the inferred type causes problems at test time, when the type must be sampled randomly without foresight of future events.
For example, if the type contains information about traffic light timings, 
the actual timing on the test data will almost surely differ from the randomly sampled one, resulting in out-of-distribution inputs to the policy which never encountered such a mismatch during training where the type was always inferred from the actual future. 
Furthermore, the agent might have learned to ignore the actual traffic light, instead relying entirely on the inferred type that was always optimal during training.
This could cause it to enter the intersection too early, resulting in potentially catastrophic consequences such as collisions.

Existing hierarchical work either assumes no external stochasticity in the environment \citep{wang2017robust,suo2021trafficsim}, relies on manually designed type representations that cannot capture external events but limit expressiveness \citep{igl2022symphony}, or relies on manually designed cost functions and type filters that mitigate the performance degradation but do not solve the underlying problem and induce biases in the learned behaviour \citep{xu2022bits}.

In this paper, we identify the challenges arising under stochastic environments and formulate them as a new form of distribution shift for hierarchical policies. 
Unlike the familiar covariate shift in the state distribution \citep{ross2011reduction}, this \emph{conditional type shift} occurs in the distribution of the inferred latent type.
It greatly reduces performance by yielding causally confused agents that rely on the latent type for information about external factors, instead of inferring them from the latest environment observation.
We propose \emph{\fullname} (\name) to eliminate this distribution shift through a coupled adversarial training objective under randomly sampled types.
We do not require access to an expert, counterfactuals, or manually specified type labels for trajectories.

Experimentally, we show the need for improved distributional realism in state-of-the-art imitation learning techniques such as GAIL \citep{ho2016generative}. Furthermore, we show that naively trained hierarchical models with inferred types improve distributional realism, but exhibit poor task performance in stochastic environments. 
By contrast, \name{} can maintain good task performance in stochastic environments while improving distributional realism.
We evaluate \name{} on the illustrative \emph{\envname{}} as well as the large scale \emph{Waymo Open Motion Dataset} \citep{ettinger2021large} of real driving behaviour.
\looseness=-1

\section{Background}
\label{sec:background}

We are given a dataset $\mathcal{D} = \{\tau_i\}_{i=1}^N$ of $N$ trajectories $\tau_i = \bs{s}_0^{(i)}, \bs{a}_0^{(i)}, \dots \bs{s}_T^{(i)}$, drawn from $p(\tau)$ of one or more experts interacting with a stochastic environment $p(\s_{t+1}|\s_t, \a_t)$ where $\bs{s}_t\in\mathcal{S}$ are states and  $\bs{a}_t\in\mathcal{A}$ are actions.
Our goal is to learn a policy $\pol(\bs{a}_t|\bs{s}_t)$ to match $p(\tau)$ when replacing the unknown expert and generating rollouts $\hat{\tau} \sim p(\hat{\tau}) = p(\bs{s}_0) \prod_{t=0}^{T-1} \pol(\ah_t|\sh_t) p(\sh_{t+1}|\sh_t, \ah_t)$ from the inital states $\s_0\sim p(\s_0)$.
We simplify notation and write $\hat{\tau} \sim \pol(\tauh)$ and $\tau \sim \mathcal{D}(\tau)$ to indicate rollouts generated by the policy or drawn from the data respectively.
Expectations $\bbE_{\tau\sim\mathcal{D}}$ and $\bbE_{\hat{\tau}\sim\pol}$ are taken over all pairs $(\s_t, \a_t) \in\tau$ and $(\sh_t, \ah_t) \in\tauh$.
\looseness=-1

Previous work \citep[e.g.,][]{ross2011reduction,ho2016generative} shows that a core challenge of learning from demonstration is reducing or eliminating the covariate shift in the state-visitation frequencies $p(\s)$ caused by accumulating errors when using $\pol$.
Unfortunately, \emph{Behavioural Cloning} (BC), a simple supervised training objective optimising $\max_{\bs{\theta}} \mathbb{E}_{\tau\sim\mathcal{D}}\left[\log\pol(\bs{a}_t|\bs{s}_t)\right]$ is not robust to it.
To overcome covariate shift, generative adversarial imitation learning (GAIL) \citep{ho2016generative} optimises $\pol$ to fool a learned discriminator $\Dash$ that is trained to distinguish between trajectories in $\mathcal{D}$ and those generated by $\pol$:
\begin{align}
\label{eq:mgail}
  \min_{\bs{\theta}} \max_{\bs{\phi}} 
  & \bb{E}_{\hat{\tau}\sim \pol}\Big[\log(\Dash)\Big]  + \\
  & \bb{E}_{\tau\sim\mathcal{D}}\Big[\log(1-\Das)\Big].
\end{align}
The policy can be optimised using reinforcement learning, by treating the log-discriminator scores as costs, $r_t = - \log \Dash$.
Alternatively, if  the policy can be reparameterized \citep{kingma2013auto} and the environment is differentiable, the sum of log discriminator scores can be optimised directly without relying on high-variance score function estimators by backpropagating through the transition dynamics, $\Ladv(\tauh) = \bbE_{\hat{\tau}\sim \pol} \left[\sum_t -\log \Dash\right]$.
We refer to this as \emph{Model-based GAIL} (MGAIL) and assume a known differentiable environment instead of a learned model as in \citep{baram2016model}.

In this work, we are concerned with multimodal distributions $p(\tau)$ and how mode collapse can be avoided when learning $\pol$.
To this end, we assume the dataset is sampled from 
%
$p(\tau) =  p(\s_0) \int p(\g)p(\x)\prod_{t=0}^T p(\a_t|\s_t, \g) p(\s_{t+1}|\s_t,\a_t,\x)d\x d\g$,
%
where $\g$ is the agent \emph{type}, expressing agent characteristics such as persona, goal, or, strategy, and $\x$ is a random variable capturing the stochasticity in the environment, i.e., $p(\s_{t+1}|\s_t,\a_t,\x)$ is a delta distribution $\delta_{f(\s_t, \a_t, \x)}(s_{t+1})$ for some transition function $f$. 
We call an agent realistic if its generated trajectories $\tauh\sim\pol$ lie in the support of $p(\tau)$. 
We call an agent \emph{distributionally realistic} if its distribution over trajectories matches the data, i.e. $p(\tauh)\approx p(\tau)$.
As we show in \cref{sec:experiments}, current non-hierarchical adversarial methods \citep{ho2016generative} are not distributionally realistic.

To combat mode collapse, hierarchical methods \citep[e.g.,][]{wang2017robust,lynch2020learning,suo2021trafficsim,igl2022symphony,xu2022bits} often rely on an encoder to infer latent agent types $\gh_e$ from trajectories during training, $\bs{\hat{g}_e} \sim \e(\gh_e|\tau)$, and optimise the control policy $\pol(\ah_{t}|\sh_{t}, \gh_e)$ to generate trajectories $\tauh_e$ similar to $\tau$: 
$\tauh_e\sim p(\tauh_e|\gh_e) = p(\s_0) \prod_{t=0}^{T-1} \pol(\ah_{t}|\sh_{t}, \gh_e) p(\sh_{t+1}|\ah_{t}, \sh_{t})$.
As ground truth trajectories are not accessible during testing, a prior $\p(\gh_p)$, which has been trained to match the marginal distribution $p_e(\gh_e) = \mathbb{E}_{\tau}\left[ \e(\gh_e|\tau)\right]$, is used to sample distributionally realistic types $\gh_p$. 
We indicate by subscript $\gh_p$ or $\gh_e$ whether the inferred type and trajectory are drawn from the prior distribution $\p(\gh_p)$ or encoder $\e(\gh_e|\tau)$. 
Subscripts are omitted for states and actions to simplify notation. 
Inferred types and predicted trajectories without subscripts indicate that either sampling distribution could be used.
For information theoretic quantities we use capital letters $\S, \A, \Ah, \Gh$ and $\X$ to denote the random variables for values $\s, \a, \ah, \gh$ and $\x$. 

\section{Conditional Type Shift}
\label{sec:problem}

\begin{figure}[t]
    \centering
     \begin{subfigure}[b]{0.58\linewidth}
         \includegraphics[width=0.9\textwidth,left]{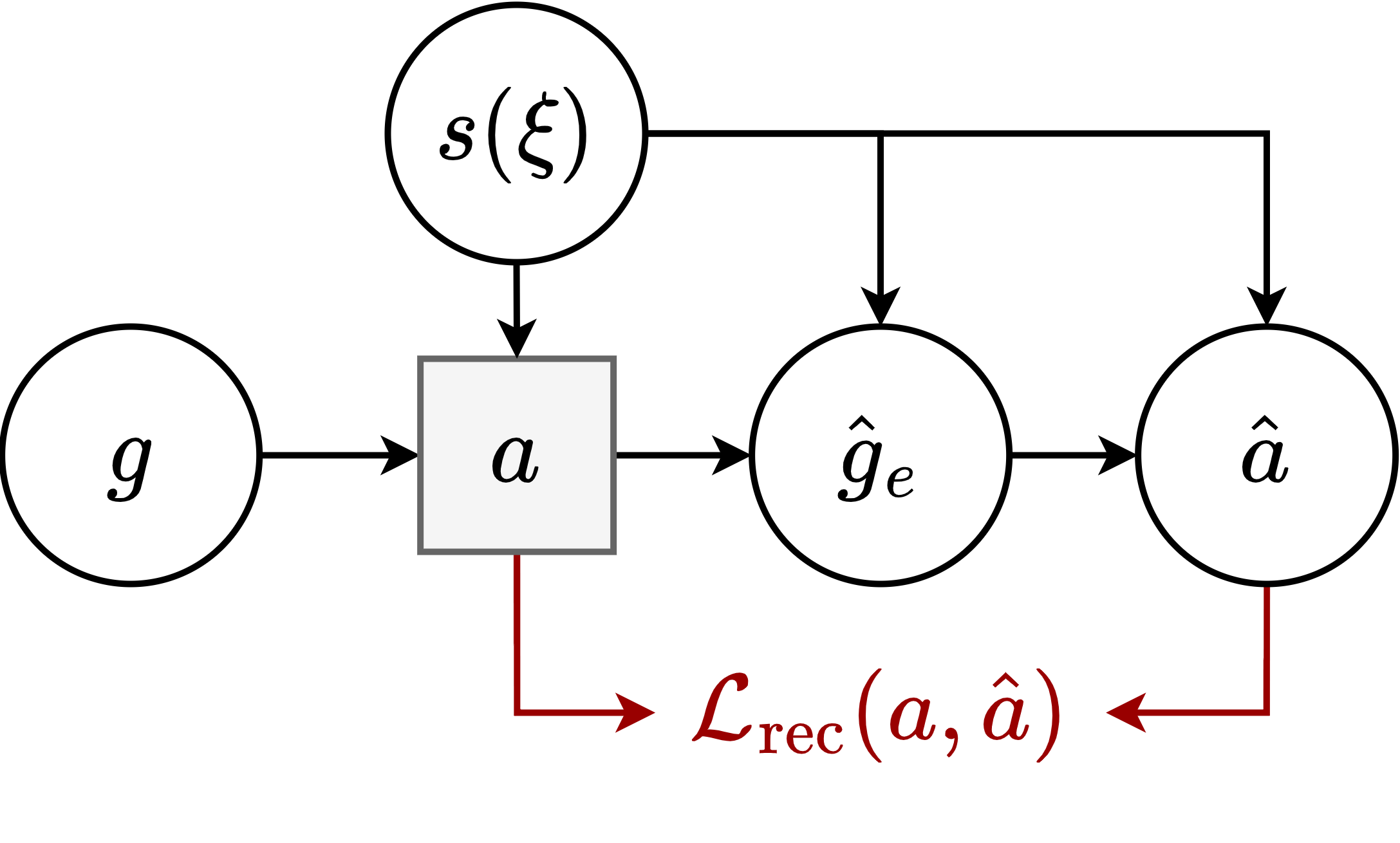}
         \caption{Encoder $\gh_e\sim\e(\gh_e|\s, \a)$ and \\ policy $\pol(\ah|\s, \gh_e)$}
         \label{fig:problem_encoder}
     \end{subfigure}
     \begin{subfigure}[b]{0.40\linewidth}
         \includegraphics[width=0.6\textwidth,center]{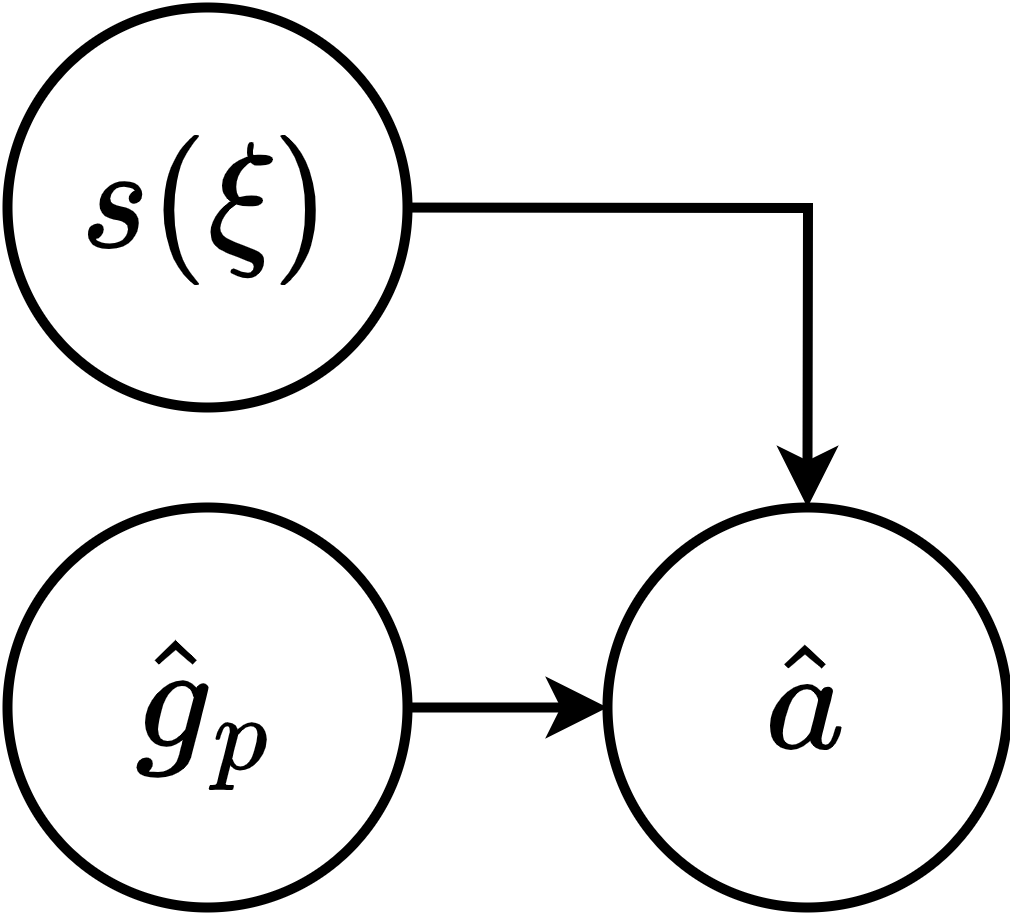}
         \vspace{1.2em}
         \caption{Prior $\gh_p\sim\p(\gh_p)$ and policy $\pol(\ah|\s, \gh_p)$.}
         \label{fig:problem_prior}
     \end{subfigure}
     \begin{subfigure}[b]{0.40\textwidth}
        \centering
         \includegraphics[width=1.0\textwidth,right]{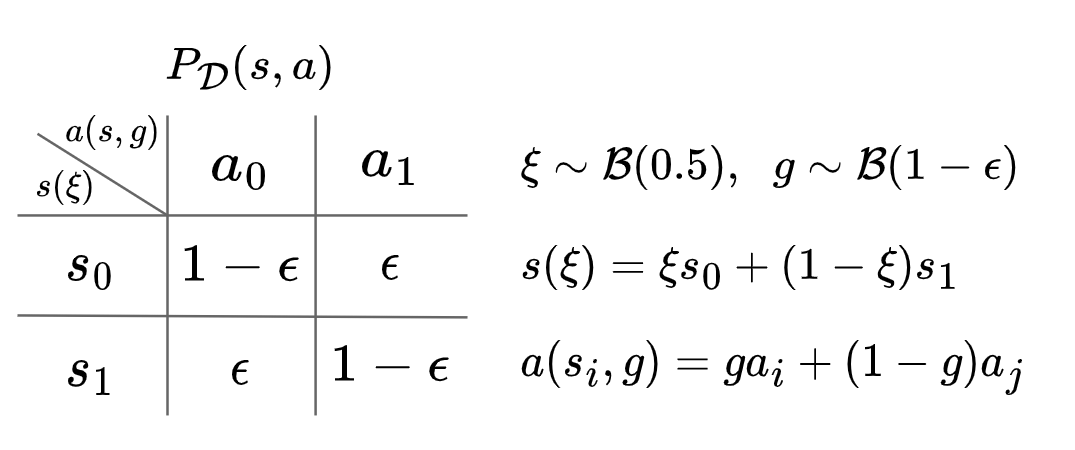}
         \caption{Example dataset.}
         \label{fig:problem_toy}
     \end{subfigure}
    \vspace{-0.5em}
    \caption{Simplified, non-temporal setup with environmental noise $\x$ and unobserved true agent type $\g$.
    The inferred type $\gh$ is sampled from $\e(\gh_e|\s, \a)$ during training (\emph{top-left}) and $\p(\gh_p)$ otherwise (\emph{top-right}).  
    The control policy is $\pol(\ah|\s, \gh)$. 
    Circles are random variables and squares deterministic functions.
    The loss $\mathcal{L}(\a, \ah)$ penalises differences between $\a$ and $\ah$.
    \emph{Bottom:} Example data, $\mathcal{B}$ denotes Bernoulli distributions.
    }
    \label{fig:problem}
    \vspace{-1.0em}
\end{figure}

Here we outline the challenge of \emph{conditional type shift} that arises for hierarchical policies in stochastic environments.
We provide a simple example illustrating the challenge and how it can be overcome, as well as formulate a proof for the exact conditions under which such a distribution shift occurs. 
These insights motivate the algorithm in \cref{sec:method}.

\subsection{Simplified model}

We use the simplified model in \cref{fig:problem}. 
For intuition, we connect it to the example mentioned in the introduction of an agent approaching a traffic light.
This model has two sources of randomness in the training data $\mathcal{D}$:
the environmental noise $\x$ (whether the traffic light is red or green) and the type $\g$ of the expert we are mimicking (whether the expert is paying attention). 
The crucial difference between $\g$ and $\x$ is that $\x$ represents \emph{external} factors outside the agent's control to which it must react, while $\g$ encodes agent-\emph{internal} decisions that can be taken independently of $\x$.
In this simple model, the temporal dimension is removed and the state $\s$ is a deterministic function of only $\x$ and not influenced by $\g$.
Hence, in this section we use $\s$ and $\x$ interchangeably.

During training, the inferred type $\gh_e$ is drawn from the encoder $\e(\gh_e|\tau)$ which has access to the future `trajectories' $\tau = (\s, \a)$ in the data.
During testing, without access to $\tau$, a prior $\p(\gh_p)$ is used to sample $\gh_p$.
Actions $\ah$ are drawn from the learned control policy $\pol(\ah|\s, \gh)$ and a reconstruction loss $\Lrec(\a, \ah)$ is minimised. 
As typical in autoencoders, the prior $\p(\gh_p)$ is trained to match the \emph{marginal} distribution of the encoder $p_e(\gh_e) = \mathbb{E}_{\tau}\left[ \e(\gh_e|\tau)\right]$ by minimizing $\Lkl(\tau) = \bbE_{\tau\sim\mathcal{D}} \left[\KL\left[\p(\gh_p)\|\e(\gh_e|\tau)\right]\right]$.

\subsection{Only external factors of influence}

We first describe a scenario with only external sources of stochasticity that serves as a minimal example of how things can go wrong due to conditional type shift.
As there are no agent-internal decisions, hierarchies are unnecessary in this minimal scenario.
In \cref{sec:problem:full}, we extend this example to include agent-internal decisions which hierarchical policies capture well. 

Consider the example data in \cref{fig:problem_toy} with $\epsilon=0$, i.e., for now we assume the expert is always paying attention.
The environment can be in two states.
Half the time, it is in $\s_0$, where the traffic light is red and the agent always takes action $\a_0=\mathrm{stop}$.
Otherwise, in $\s_1$, the traffic light is green and the agent takes action $\a_1=\mathrm{go}$.

During training, the encoder observes the actual future $\tau = (\s, \a)$ in the data and proposes the type $\gh_e\sim\e(\gh_e|\tau)$ with $\gh_e \in \{0, 1\}$. This allows, for example, the following \textbf{solution 1} which minimises the reconstruction loss $\Lrec(\a, \ah)$:
\begin{equation*}
\gh_e(\s_i, \a_j) = j \quad \text{and} \quad \pol(\ah|\s_i, \gh_e) = \a_{\gh_e} 
\end{equation*}
The encoder encodes the desired action in the type $\gh_e$ and the policy follows $\gh_e$ while ignoring $\s$.
This constitutes a perfect solution during training. 
However, during testing, we do not have access to $\tau$ and must instead draw types randomly from the prior $\gh_p\sim\p(\gh_p)$, which matches the \emph{marginal} distribution of the encoder, i.e., $\p(\gh_p) = p_e(\gh_e) = \mathcal{B}(0.5)$. 
Here, $\mathcal{B}$ is the Bernoulli distribution and the prior is drawing types $\gh_p \in \{0, 1\}$ with equal probability because it cannot know the stochastic environment state $\s$ in advance.

The conditional type shift arises because this marginal distribution does not need to match the \emph{conditional} type distribution in specific states, i.e., $\p(\gh_p) \neq \e(\gh_e|\s, \a)$. 
For example, the prior might sample $\gh_p=1$ while the stochastic environment shows a red light ($\s=\s_0$).
The resulting input to the policy, $(\s_0, \gh=1)$, was never seen during training where state and type always matched, i.e., the input pairs were either $(\s_0, \gh=0)$ or $(\s_1, \gh=1)$.

If the policy generalises to this new input by following the type, as was optimal during training, it randomly $\mathrm{stop}$s or $\mathrm{go}$es, clearly not reproducing the data distribution and causing potentially catastrophic mistakes such as collisions.

This problem always occurs when information about \emph{external} stochastic factors is captured by the type.
As there are no internal decision by the agent in this example, the ideal solution is for the type to not encode any information.
In \cref{sec:problem:full} we show that the conditional type shift problem does not arise when only agent-internal decisions are encoded, as these can be taken by the agent independently from the environment stochasticity.

\subsection{External and internal factors of influence} 
\label{sec:problem:full}

To express this, we now introduce $\epsilon > 0$ as the probability that the agent decides not to pay attention to the traffic light.
Hence, the expert now either follows the traffic light with $p(\a_i|\s_i) = 1 - \epsilon$, or ignores it with $p(\a_{\neq i}|\s_i) = \epsilon$.

The previous \emph{solution 1} is still viable during training, minimising the reconstruction loss, but still fails during testing as it generates an action distribution which ignores the traffic light 50\% of the time, i.e., $p(\ah_{\neq i}|\s_i) = 0.5$, in contrast to the expert, which only deviates with probability $p(\a_{\neq i}|\s_i) = \epsilon$.

By contrast, \textbf{solution 2} avoids the conditional type shift but successfully encodes the agent-internal decision:
\begin{equation*}
\gh_e(\s_i, \a_j) = \begin{cases}
    0 & \!\!\text{if } i=j   \\
    1 & \!\!\text{if } i\ne j
\end{cases},
\;
\pol(\ah|\s_i, \gh) = \begin{cases}
a_i & \!\!\text{if } \gh=0 \\
a_{\neq i} & \!\!\text{if } \gh=1
\end{cases}
\end{equation*}
Here the latent type $\gh$ only captures whether the agent pays attention ($\gh=0$) or not ($\gh=1$).
Now the marginal encoder type distribution is $p_e(\gh_e=0)=1-\epsilon$ and hence we have for the learned prior $\p(\gh_p)= \mathcal{B}(\epsilon)$, correctly reproducing the data in all states.

To summarize, hierarchical policies in stochastic environments only generalise at test time when the type only conveys information about agent-internal features and is uncorrelated with any external stochastic events during training.

For simplicity, the model in \cref{fig:problem_toy} has only one time-step.
For temporally extended data, the states $\s_t$ depend not only on $\x$, but also on $\g$ or $\gh$, complicating theoretical treatment. 
Nevertheless, seeing $\x$ as all \emph{future} stochasticity in the environment, the same challenges arise.
In realistic driving scenarios $\x$ not only captures traffic lights, but also the reactions of other agents in the scene. 
Similarly, agent-internal factors include a wide range of information such as goals, driving-style or level of attention.

\subsection{Theorem}

Here we provide an information theoretic proof that agents that, during training, rely on the inferred type to acquire information about external events, do not react appropriately to environment stochasticity during testing.
This formalises the previous discussion but is not needed to follow subsequent sections of the paper.

\textbf{Theorem 1 }
\emph{
The hierarchical autoencoding model $\p(\ah|\s, \a)$ and test policy $\p(\ah|\s)$ are as described above, sampling latent types from encoder $\e$ and prior $\p$ respectively.
We assume an optimal reconstruction loss $\Lrec=0$ on the training data $P_\mathcal{D}(\s,\a)$.
For the training distribution $P(\s, \a, \gh_e)=P_\mathcal{D}(\s, \a)\e(\gh_e|\s, \a)$ and test-time distribution $P(\s, \ah, \gh_p)=P_\mathcal{D}(\s)\p(\gh_p)\pol(\ah|\s, \gh_p)$ we have that if $H(\A|\Gh_e) < I(\S, \A) $ and $H(\A|\Gh_e) = H(\Ah|\Gh_p)$, then  $I(\S, \Ah) < I(\S, \A)$ and consequently $H(\Ah|S) > H(\A|S)$.
}

We denote by $H(X)$ the entropy, by $H(X|Y)$ the conditional entropy, by $I(X, Y)$ the mutual information and by $I(X, Y| Z)$ the conditional mutual information between random variables.
Intuitively, $H(\A|\Gh_e) < I(\S, \A)$ if the encoder captures information about $\A$ in $\Gh_e$ that is also accessible through $\S$, i.e., information about external events.
The condition $H(\A|\Gh_e) = H(\Ah|\Gh_p)$ implies that the policy relies on this information in $\Gh$ to predict $\Ah$.
If both conditions are true then $H(\Ah|S) > H(\A|S)$, stating that the state $\S$ has less predictive power for the predicted action $\Ah$ than for actions $\A$ in the dataset, i.e.: that the policy is ignoring action-relevant information in the states.

\textbf{Proof }
The proof relies on the \emph{interaction information} $I(X, Y, Z)$, an extension of mutual information to three variables.
Importantly, the interaction information can be positive or negative. A positive interaction information indicates that one variable explains some of the correlation between the other two while a negative interaction information indicates that one variable enhances their correlation.

The model $\p(\ah|\s, \a)=\e(\gh_e|\a, \s)\pol(\ah|\s, \gh_e)$ is trained on the dataset $P_\mathcal{D}(\s, \a)$. 
Achieving minimal reconstruction loss is achieved only when $\ah=\a$ is predicted with certainty, implying $H(\Ah|\S, \Gh_e)=H(\Ah|\S, \Gh_p)=0$.

During training on the dataset $P_\mathcal{D}(\s, \a)$ the interaction information is positive because $H(\A|\Gh_e) < I(\S, \A)$:
\begin{align*}
    I(\A, \Gh_e, \S) = &  I(\S, \A) - I(\S, \A | \Gh_e) = \\
                       & I(\S, \A) - H(\A|\Gh_e) + H(\A |\S, \Gh_e) > 0.
\end{align*}

On the other hand, during testing, we have $I(\Gh_p, \S) = 0$ because $\Gh_p$ is drawn independently of $\S$. 
The interaction information becomes weakly negative:
\begin{equation*}
    I(\Ah, \Gh_p, \S) = I(\Gh_p, \S) - I(\Gh_p, \S | \Ah) \le 0
\end{equation*}

With $I(\Ah, \Gh_p, \S) = I(\S, \Ah) - H(\Ah | \Gh_p)$ we get
\begin{equation}
    \label{eq:proof_2}
    I(\S, \Ah) - H(\Ah | \Gh_p) \le 0 < I(\S, \A) - H(\A|\Gh_e)
\end{equation}
and hence, because by assumption $H(\A|\Gh_e) = H(\Ah|\Gh_p)$, this gives us the desired result $I(\S, \Ah) < I(\S, \A)$ from which $H(\Ah|S) > H(\A|S)$ follows directly.

\section{Robust Type Conditioning}
\label{sec:method}

\begin{figure*}[t]
\includegraphics[width=0.80\textwidth]{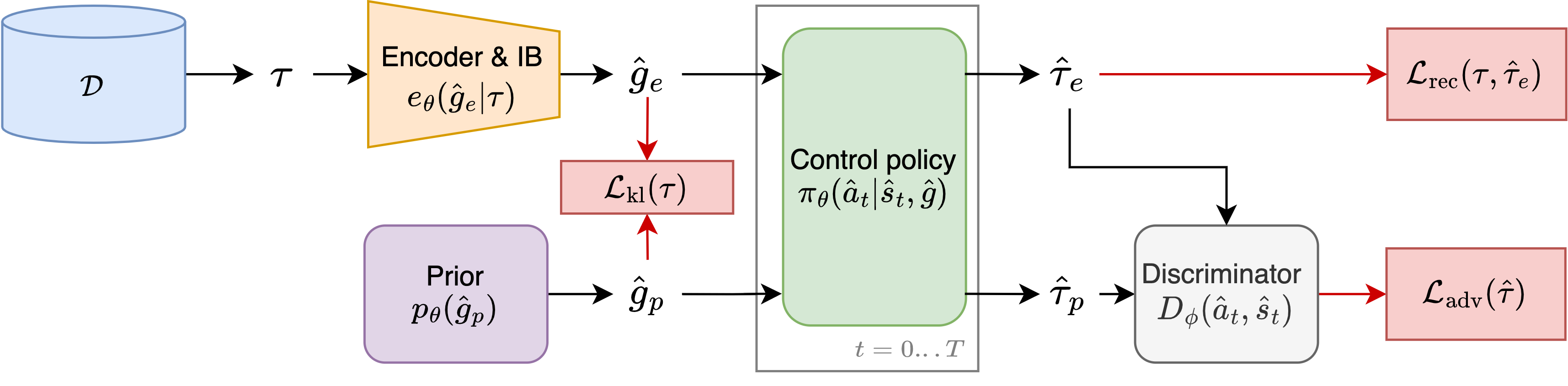}
\centering
\caption{\emph{\fullname{}} (\name{}): The control policy $\pol(\ah_t|\sh_t, \gh)$ is trained under inferred types $\gh$ sampled from both the encoder $\e(\gh_e|\tau)$ and the prior $\p(\gh_p)$. 
The hierarchical loss $\Lvae(\tau, \tauh) = \Lrec(\tau, \tauh_e) + \beta \Lkl(\tau)$ improves distributional realism.
The adversarial loss $\Ladv(\tauh)$ under prior types prevents causally confused policies and ensures good task performance at test time, even in stochastic environments.
$\Lkl(\tau)$ optimises the prior to sample distributionally realistic types.
\looseness=-1
}
\vspace{-1.5em}
\label{fig:architecture}
\end{figure*}

We present \gls{rtc}, a method for improving distributional realism in imitation learning while maintaining high task performance, even in stochastic environments.
As shown in previous work \citep{suo2021trafficsim,igl2022symphony,xu2022bits}, and  confirmed in \cref{sec:experiments}, hierarchical policies trained in an autoencoder framework are currently the most effective approach at improving distributional realism.
However, such policies require the latent type to be inferred from the future trajectory, which can cause problems in stochastic environments (see \cref{sec:problem}).

To overcome this limitation, we propose to combine the autoencoder training objective $\Lvae = \Lrec + \beta \Lkl$ with an additional adversarial objective $\Ladv$ utilising a learned discriminator $\Das$. 
Importantly, this additional objective allows us to sample training types not only from the encoder, but also from the prior.
When types are sampled from the prior, the hierarchical loss $\Lvae$ cannot be used as we generally do not have access to ground truth trajectories corresponding to this specific type, which are required for the reconstruction loss $\Lrec$.
Instead, for these types, we only optimise the adversarial objective $\Ladv(\tauh)=\sum_t -\log \Dash$.

During training, we hence split each minibatch $\mathcal{B}=\{\tau^{(b)}\}_{b}^{N_b}$ of $N_b$ trajectories sampled from $\mathcal{D}$ into two parts. 
For the fraction $f$ of trajectories in $\mc{B}{}$ the rollouts $\tauh_e$ are generated from types sampled from the encoder $\gh_e\sim\e(\gh_e|\tau)$ and objectives $\Ladv + \lambda \Lvae$ are optimised (first line in \cref{eq:ric}).
For the remaining fraction $(1-f)$ of trajectories, types are sampled from the prior $\p(\gh_p)$ and only $\Ladv$ is optimised (second line in \cref{eq:ric}).

Because the policy does not know whether the type is sampled from the encoder or prior, this combination ensures that policies follow agent-internal information in the type, due to the autoencoder training objective, but ignore any information in the type about external stochastic events, as this would lead to unrealistic trajectories under prior types, which are penalised by the adversarial training objective. 
Lastly, because the KL objective $\Lkl(\tau) = \bbE_{\tau\sim\mathcal{D}} \left[\KL\left[\p(\gh_p)\|\e(\gh_e|\tau)\right]\right]$ minimises the amount of information encoded in the type, such unused information would not even be encoded.

Consequently, the full \gls{rtc} loss is 
\begin{equation}
\label{eq:ric}
    \begin{alignedat}{1}
    \mc{L}{\name} = \bbE_{\mc{D}{}(\tau)\e(\gh_e|\tau)\pol(\tauh_e|\gh_e)} & \big[\lambda \Ladv(\tauh) + \Lvae(\tau, \tauh) \big]  \\ 
  + \; \bbE_{\mc{D}{}(\tau)\pt(\gh_p)\pol(\tauh_p|\gh_p)} & \big[ \lambda\Ladv(\tauh)\big],
    \end{alignedat}
\end{equation}
with 
\begin{align*}
\Lvae(\tau, \tauh) & = \Lrec(\tau, \tauh) + \beta \Lkl(\tau) \\
\Lkl(\tau) & = \bbE_{\tau\sim\mathcal{D}} \left[\KL\left[\p(\gh_p)\|\e(\gh_e|\tau)\right]\right] \\
\Ladv(\tauh) & = \sum_t -\log \Dash,
\end{align*}
where $\p(\gh_p)$ is a learned prior, $\e(\gh_e|\tau)$ a learned trajectory encoder and $\pol(\tauh|\gh)$ is shorthand for generating trajectories $\tauh$ by rolling out the learned control policy $\pol(\ah|\sh, \gh)$ in the environment.
Parameters $\bs{\bar{\theta}}$ are held fixed and $\lambda$ and $\beta$ are scalar weights.
$\Das$ is a learned per-timestep discriminator.
Lastly, $\Lrec(\tau, \tauh)$ is a reconstruction loss between $\tau$ and $\tauh_e$ which can take different forms.
For example, in \cref{sec:toyexp} we use the BC loss $\Lrec(\tau, \tauh) = -\log\pol(\a_t|\s_t, \gh_e)$
while in \cref{sec:womd} we minimise the $L_2$ distance between agent positions in $\s_t$ and $\sh_{t}$.
The loss $\Lkl(\tau)$ optimises the prior to propose \emph{distributionally realistic} types by matching the marginal encoder distribution.

One can understand the problem of conditional type shift as one of \emph{causally confused} policies which refer to the type for information about external stochastic events, instead of acquiring this information directly from the currently observed states.
From this perspective, sampling from the prior constitutes a causal intervention $do(\gh)$ in which $\gh$ is changed independently of the environmental factor $\x$.
\citep{de2019causal} show that causal confusion can be avoided by applying such interventions and optimising the policy to correctly predict the counterfactual expert trajectory distribution, in our case $p_{\text{expert}}(\tau | \x, do(\gh))$.
Unfortunately, we do not have access to this counterfactual trajectory.
Instead, we rely on the generalisation of $\pol$ to get us `close' to such a counterfactual trajectory for types $do(\gh)$ and then refine the policy locally using the adversarial objective.

We find that both continuous type representations with and discrete type representations using straight-through gradient estimation work well in practice (see \cref{sec:womd}).

Optimisation of $\Ladv$ and $\Lrec$ can either be performed directly, similar to MGAIL 
\citep{baram2016model}, by using a differentiable environment and reparameterised policies and encoder \citep{kingma2013auto} or by treating them as rewards and using RL methods such as TRPO \citep{schulman2015trust,ho2016generative} or PPO \citep{schulman2017proximal}. 
The loss $\Lkl$ can always be optimised directly.

\section{Related Work}
\label{sec:related}

Hierarchical policies have been extensively studied in RL \citep[e.g.,][]{sutton1999between,bacon2017option,vezhnevets2017feudal,nachum2018nearoptimal,igl2020multitask} and IL.
In RL, they improve exploration, sample efficiency and fast adaptation.
By contrast, in IL, hierarchies are used to capture multimodal distributions, improve data efficiency \citep{krishnan2017ddco,le2018hierarchical}, and enable goal conditioning \citep{shiarlis2018taco}.
Similar to our work, \citep{wang2017robust} and \citep{lynch2020learning} learn to encode trajectories into latent types that influence a control policy. 
Crucially, both only consider deterministic environments and hence avoid the distribution shifts and unwanted information leakage we address. 
They extend prior work in which the type, or context, is provided in the dataset \citep{merel2017learning}, which is also assumed in \citep{fei2020triple}.
\citep{tamar2018imitation} use a sampling method to infer latent types.
\citep{khandelwal2020if} and \citep{igl2022symphony} use manually designed encoders specific to road users by expressing future goals as sequences of lane segments.
This avoids information leakage but cannot express all characteristics of human drivers, such as persona, and cannot transfer to other tasks.
BITS \citep{xu2022bits} uses goal positions as types, which suffer from conditional type shift. 
Consequently, their method requires behaviour prediction and a manually specified cost function to filter goals that might mismatch with predicted futures.
Futures states in deterministic environments \citep{ding2019goal}, language \citep{pashevich2021episodic}, and predefined strategy statistics \citep{vinyals2019grandmaster} have also been used as types.
\looseness=-1

Information theoretic regularization offers an alternative to learning hierarchical policies using the auto-encoder framework \citep{li2017infogail,NIPS2017_632cee94}. 
However, these methods are less expressive since their prior distribution cannot be learned and only aim to cluster modes already captured by the agent but not penalize dropping modes in the data.
This provides a useful inductive bias but often struggles in complex environments with high diversity, requiring manual feature engineering \citep{eysenbach2018diversity,pathak2019self}. 
\looseness=-1

Lastly, TrafficSim \citep{suo2021trafficsim} uses IL to model driving agents, controlling all stochasticity in the scene but using independent prior distributions for separate agents.
Hence, while the environment is assumed deterministic, conditional type shift can occur between the separate agent-types which are correlated during training but independent during testing.
They use a biased ``common sense'' collision avoidance loss, motivated by covariate shift in visited states. 
Our work suggests that type shift might also explain the benefits gained. 
In contrast, our adversarial objective is unbiased.

\section{Experiments}
\label{sec:experiments}

\begin{figure}
     \centering
     \begin{subfigure}[b]{0.3\linewidth}
         \centering
         \includegraphics[width=\textwidth]{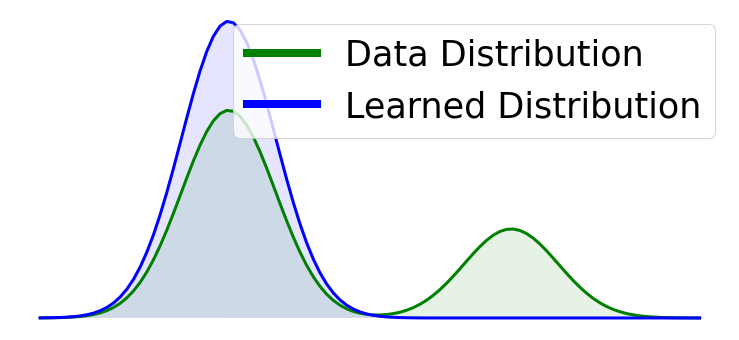}
         \caption{}
         \label{fig:dr_1}
     \end{subfigure}
     \hfill
     \begin{subfigure}[b]{0.3\linewidth}
         \centering
         \includegraphics[width=\textwidth]{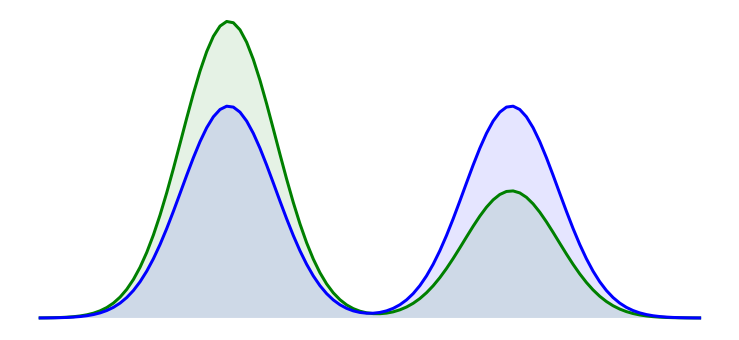}
         \caption{}
         \label{fig:dr_2}
     \end{subfigure}
     \hfill
     \begin{subfigure}[b]{0.3\linewidth}
         \centering
         \includegraphics[width=\textwidth]{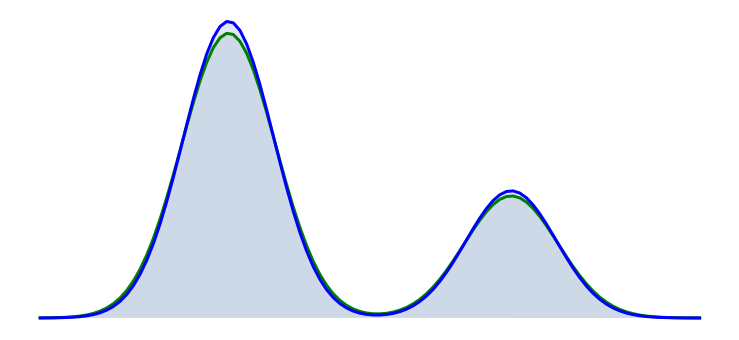}
         \caption{}
         \label{fig:dr_3}
     \end{subfigure}
        \caption{Differences between \emph{realism}, \emph{coverage} and \emph{distributional realism}. The data distribution $P(X_{D})$ is shown in green, blue denotes a learned distribution $P_\theta(X_{L})$.
        \textbf{(a)} Data from the learned distribution is \emph{realistic}, i.e. $\supp(X_{L}) \subseteq \supp(X_{D})$, but not \emph{distributionally realistic.}
        \textbf{(b)} The learned distribution achieves \emph{coverage} but not \emph{distributional realism}: the frequencies of modes are not matched.
        \textbf{(c)} The learned distribution is \emph{distributionally realistic}.
        In practice we measure distributional realism in selected features $h(X)$ as the dimensionality of $\mathcal{X}$ is too high.
        }
        \label{fig:distr_realism_example}
        \vspace{-1.5em}
\end{figure}

\begin{table*}[t]
\captionsetup{justification=centering}
\centering
\caption{Averages and standard deviation over 20 training runs on \emph{WOMD}. The best two values are highlighted. }
\label{tab:womd}
\renewcommand{\arraystretch}{1.2}
\resizebox{0.75\textwidth}{!}{
\begin{tabular}{l|ccccc}
\Xhline{2\arrayrulewidth}
         & \begin{tabular}[c]{@{}c@{}}Collision\\ rate (\%) $\downarrow$ \end{tabular} 
         & \begin{tabular}[c]{@{}c@{}}Off-road\\ time (\%) $\downarrow$ \end{tabular}  
         & \begin{tabular}[c]{@{}c@{}}MinADE\\ (m) $\downarrow$\end{tabular} 
         & \begin{tabular}[c]{@{}c@{}}Curvature JSD\\($\times10^{-3}$) $\downarrow$\end{tabular}  
         & \begin{tabular}[c]{@{}c@{}}Progress JSD\\ ($\times10^{-3}$) $\downarrow$\end{tabular} \\ 
\hline

Data Distribution & 1.16  & 0.68 & - & - & - \\ 

\hline

MGAIL            &        5.39 $\pm$ 0.68 &        0.89 $\pm$ 0.12 &        1.34 $\pm$ 0.08 &        1.32 $\pm$ 1.48 &        3.81 $\pm$ 1.29 \\
Symphony         &        6.39 $\pm$ 0.95 &        0.90 $\pm$ 0.06 &        1.40 $\pm$ 0.12 &        0.97 $\pm$ 0.62 &        6.44 $\pm$ 5.25 \\
InfoMGAIL - C    &        5.21 $\pm$ 0.37 &        0.89 $\pm$ 0.14 &        1.29 $\pm$ 0.07 &        1.24 $\pm$ 0.93 &        4.40 $\pm$ 1.47 \\
InfoMGAIL - D  &        4.82 $\pm$ 0.29 &        0.84 $\pm$ 0.10 &        1.35 $\pm$ 0.11 &  {\bf 0.77 $\pm$ 0.44} &        4.01 $\pm$ 1.45 \\
NaiveHierarchy &       35.08 $\pm$ 0.44 &        1.83 $\pm$ 0.42 &        {\bf1.12 $\pm$ 0.01 }&        1.76 $\pm$ 2.05 &        {\bf 2.54 $\pm$ 0.63} \\
RTC - C       &     {\bf  4.23 $\pm$ 0.16} &  {\bf 0.68 $\pm$ 0.04} &        1.15 $\pm$ 0.10 &  {\bf 0.43 $\pm$ 0.06} &  {\bf 2.17 $\pm$ 0.65} \\
RTC - D          &  {\bf 4.21 $\pm$ 0.24} &  {\bf 0.74 $\pm$ 0.06} &        {\bf1.12 $\pm$ 0.10 }&        0.89 $\pm$ 0.66 &        2.56 $\pm$ 0.54 \\

\hline

\Xhline{2\arrayrulewidth}
\end{tabular}
}
\vspace{-1em}
\end{table*}


We show in two stochastic environments with multimodal expert behaviour that i) existing \gls{il} methods suffer from insufficient distributional realism, 
ii) hierarchical methods can suffer from conditional type shift and degrading task performance, and
iii) \name{} improves distributional realism while maintaining excellent task performance.

We compare the following models:
\MGAIL{} uses an adversarial training objective with learned discriminator. It also optimises a \BC{} loss as we found this to improve performance.
\Symphony{} \citep{igl2022symphony} (called `MGAIL+H' in the original paper), building on \MGAIL{}, utilises future lane segments as manually specified types which avoid conditional type shift but limit expressiveness.
\InfoMGAIL{} \citep{li2017infogail} augments \MGAIL{} to elicit distinct trajectories for different types by using an information-theoretic loss.
This is an alternative training paradigm for hierarchical policies, besides using autoencoders with reconstruction loss.
For fair comparison, our method $\name$ uses the same \MGAIL{} implementation as adversarial objective.
We investigate both continuous and discrete type representations, \nameC{} and \nameD{}.
\HNoPT{} is a hierarchical autoencoder not training on prior-sampled types (but also using the adversarial \MGAIL{} objective) and hence experiencing conditional type shift and high collision frequency.

\subsection{\envname{}}
\label{sec:toyexp}

\begin{figure}[ht]
    \centering
    \begin{subfigure}{0.99\linewidth}
        \includegraphics[width=\textwidth]{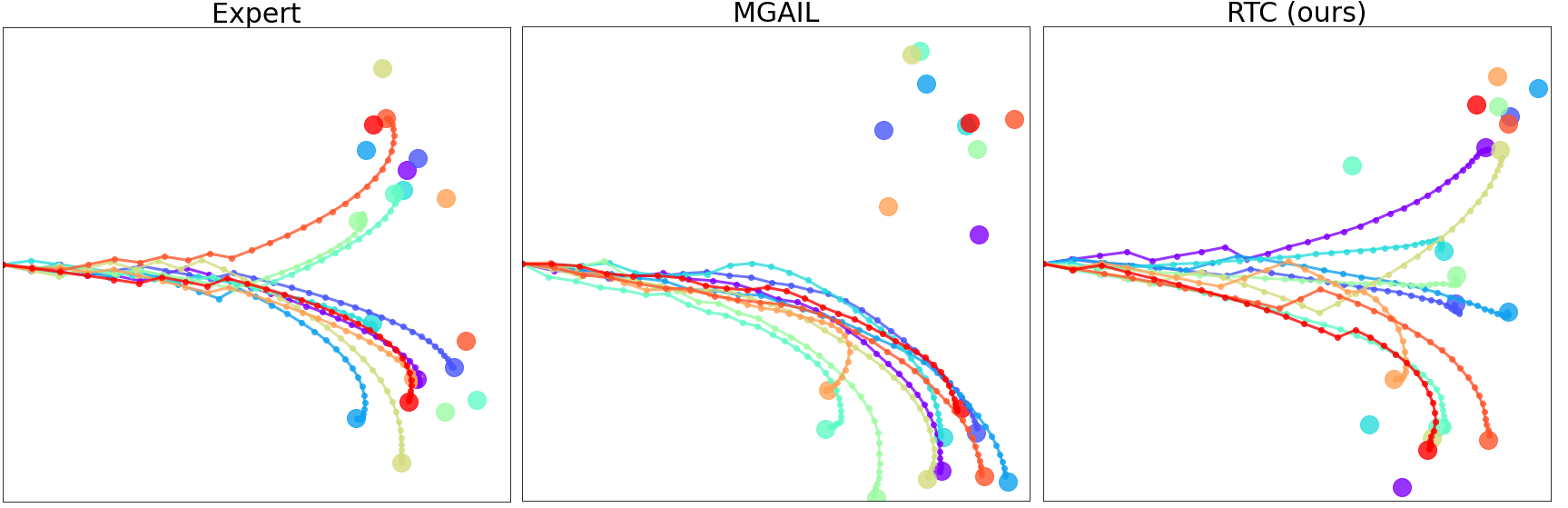}
        \centering
    \end{subfigure}
    \begin{subfigure}{0.99\linewidth}
        \includegraphics[width=\textwidth]{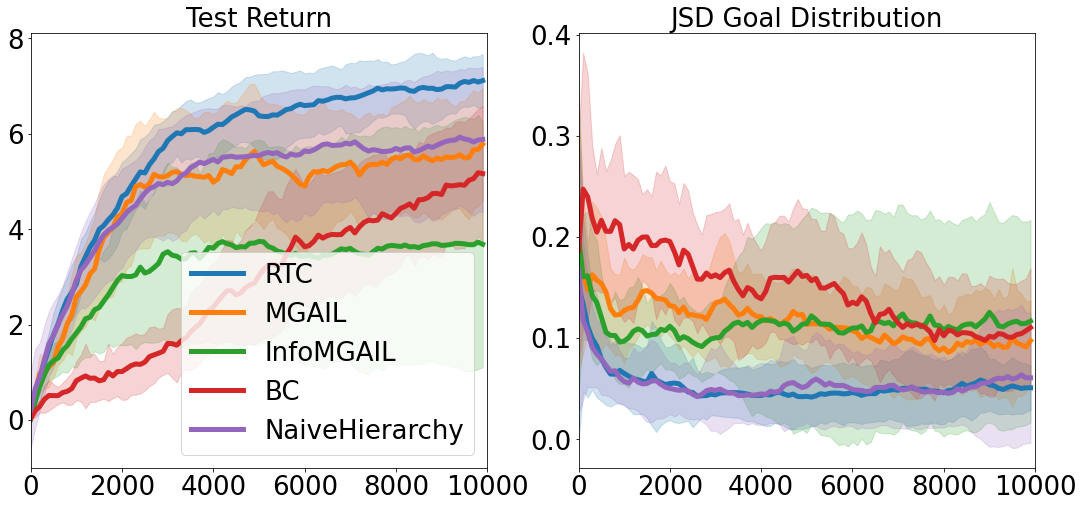}
        \centering
    \end{subfigure}
    \caption{
    \emph{Top:} Visualization of ten randomly sampled goal pairs and associated trajectories.
    \emph{Bottom:} Training curves, exponentially smoothed and averaged over 20 seeds.
    Shading shows the standard deviation.
    We show task performance as `Test Return' (higher is better) and distributional realism as `JSD' between the goal distribution of expert and agent (lower is better). 
    }
    \label{fig:toy_vis}
\end{figure}

In the double goal problem, the expert starts from the origin and creates a multimodal trajectory distribution by randomly choosing and approaching one of two possible, slowly moving goals located on the 2D plane. 
Stochasticity is introduced through randomized initial goal locations and movement directions. 
Nevertheless, the \emph{lower} and \emph{upper} goal $\{\g_l, \g_u\}$ remain identifiable by their location as $y_l < 0$ for $\g_l$ and $y_u>0$ for $\g_u$ (see \cref{fig:toy_vis}).
While both goals are equally easy to reach, the expert has a preference $P(G=\g_l)=0.75$.
Sufficiently complex expert trajectories prevent \BC{} from achieving optimal performance, requiring more advanced approaches.
The expert follows a curved path and randomly resamples the selected goal for the first ten steps to avoid a simple decision boundary along the $x$-axis in which experts in the lower half-plane always target goal $\g_l$. 
\name{} uses the BC loss as reconstruction loss $\Lrec(\tau) = -\log\pol(\a_t|\s_t, \gh_e)$ and continuous types.
All policies use a bimodal Gaussian mixture model as action distribution.

This experiment combines agent-internal decisions (which goal to approach) with external stochasticity (goal starting positions and movement directions).
Task performance is measured as the number of steps for which the agent is within $\delta=0.1$ distance of one of the goals (higher is better). 
Distributional realism is measured as the divergence between the empirical distributions, $\JSD\left(p_{\text{agent}}(h_s)\|p_{\text{expert}}(h_s)\right)$ (lower is better) where
we take $h_s=\sign(y_T)$ of the final agent position $[x_T, y_T]$ to indicate which goal was approached.
Our aim is to improve distributional realism while maintaining or improving task performance.

\Cref{fig:toy_vis} shows that \MGAIL{} improves task performance compared to \BC{}.
Our method, \name{}, improves it further, possibly because given a type, the required action distribution is unimodal.
Importantly, \name{} substantially improves distributional realism, achieving lower JSD values. 
The bias introduced by the information-theoretic loss in \InfoMGAIL{} reduces task performance without improving distributional realism.
Lastly, \HNoPT{} achieves excellent distributional realism through the learned hierarchy but suffers reduced task performance due to conditional type shift.

\subsection{Waymo Open Motion Dataset (WOMD)}
\label{sec:womd}

To evaluate \name{} on a complex environment we use the \emph{Waymo Open Motion Dataset} \citep{ettinger2021large} consisting of $487K$ segments of real world driving behaviour, each $9s$ long at $10Hz$. 
We follow \citep{igl2022symphony} by controlling agents at $3.3Hz$ and replying uncontrolled agents from logs.
Distributionally realistic agents are critical for driving simulations, for example for estimating safety metrics.
Diverse intents and driving styles cause the data to be highly multimodal.
External stochasticity is induced through the unpredictable behaviour of other cars, cyclists and pedestrians.
We use $\Lrec(\tau, \tauh)=\sum_t^T\LHuber(\s_t, \sh_t)$ where $\LHuber$ is the average Huber loss of the four vehicle bounding box corners.

The percentage of segments with collisions and time spent off-road are proxy metrics for task performance and realism.
Mode coverage is measured by the minimum average displacement error,
$\minADE = \bbE_{\tau\sim\mathcal{D}, \{\tauh_i\}_i^{K}\sim\pol} \left[\min_{\tauh_i} \frac{1}{T}\sum^{T}_{t=1}  \delta(\s_{t}, \sh_{i,t}) \right]$,
where $\delta$ is the Euclidean distance between agent positions and we find the minimum over $K=16$ rollouts (hierarchical methods use $K$ independently sampled types).
Lower \emph{minADE} implies better mode coverage, but does not directly measure the relative frequency of modes, e.g., low probability modes may be overrepresented.
To measure distribution matching in driving intent, we use the \emph{Curvature JSD} \citep{igl2022symphony}:
in lane branching regions, such as intersections, it maps trajectories to the nearest lane and extracts its curvature as feature $h_{cur}$. 
The driving style distribution is measured through the progress feature $h_{style}=\delta(\sh_0, \sh_T)$. 
To compute $\JSD\left(p_{\text{agent}}(h_{cur/style})\|p_{\text{expert}}(h_{cur/style})\right)$, the value of $h_{cur/style}$ is discretize into 100 equisized bins.

Results are provided in \cref{tab:womd}.
Both versions of \name{} improve task performance (collisions and off-road events) and distributional realism metrics (minADE and divergences) compared to the flat \MGAIL{} baseline and previous hierarchical approaches (\Symphony{}, \InfoMGAIL{}, \HNoPT).
Both type representations, \nameC{} and \nameD{}, perform similarly, showing robustness of \name{} to different implementations.

\MGAIL{} achieves good task performance, but is outperformed by \name{} due to the use of hierarchy.
On the other hand, \Symphony{}, using lane segment goals to capture driving intent, consequently improves on the \emph{Curvature JSD} distributional realism metric, but not on \emph{Progress JSD} which measures driving style, not intent.
In contrast, \name{} improves on both distributional realism metrics since the fully learned type is more expressive.
The information-theoretic loss in \InfoMGAIL{} improves distributional realism on some metrics, but is less effective than \name: while the additional \InfoMGAIL{} loss ensures that the type contains some information, it does not require this information to be useful, unlike in an autoencoder framework.

Lastly, the advantage of \name{} in achieving \emph{both} good task performance and distributional realism becomes clearest by comparing it to \HNoPT{}. 
While \HNoPT{} achieves some improvements in distributional realism, is has nearly an order of magnitude more collisions.
This is a consequence of the challenges discussed in \cref{sec:problem}: At training time, the inferred type contains too much information, for example when to break or start driving. At test time, because this information is sampled independently to what is actually happening in the environment, the agent behaves incorrectly and collides with other road users.

\section{Conclusions, Limitations, and Future Work}
\label{sec:conclusions}

This paper identified new challenges in learning hierarchical policies from demonstration to capture multimodal trajectory distributions in stochastic environments.
We expressed them as \emph{conditional type shifts} in the hierarchical policy.
We proposed \emph{\fullname{}} (\name) to eliminate these distribution shifts and showed improved distributional realism while maintaining or improving task performance on two stochastic environments, including the Waymo Open Motion Dataset \citep{ettinger2021large}.
Future work will address \emph{conditional} distributional realism by not only matching the marginal distribution $p(\tau)$, but the conditional distribution $p(\tau|\x)$ under a specific realization of the environment.
For example, drivers might change their intent based on the current traffic situation or players might adapt their strategy as the game unfolds.
Achieving such conditional distributional realism will also require new models and metrics.

{\footnotesize
\bibliography{refs}}

\begin{thebibliography}{10}
\providecommand{\url}[1]{#1}
\csname url@samestyle\endcsname
\providecommand{\newblock}{\relax}
\providecommand{\bibinfo}[2]{#2}
\providecommand{\BIBentrySTDinterwordspacing}{\spaceskip=0pt\relax}
\providecommand{\BIBentryALTinterwordstretchfactor}{4}
\providecommand{\BIBentryALTinterwordspacing}{\spaceskip=\fontdimen2\font plus
\BIBentryALTinterwordstretchfactor\fontdimen3\font minus
  \fontdimen4\font\relax}
\providecommand{\BIBforeignlanguage}[2]{{%
\expandafter\ifx\csname l@#1\endcsname\relax
\typeout{** WARNING: IEEEtran.bst: No hyphenation pattern has been}%
\typeout{** loaded for the language `#1'. Using the pattern for}%
\typeout{** the default language instead.}%
\else
\language=\csname l@#1\endcsname
\fi
#2}}
\providecommand{\BIBdecl}{\relax}
\BIBdecl

\bibitem{amodei2016concrete}
D.~Amodei, C.~Olah, J.~Steinhardt, P.~Christiano, J.~Schulman, and D.~Man{\'e},
  ``Concrete problems in ai safety,'' \emph{arXiv preprint arXiv:1606.06565},
  2016.

\bibitem{hadfield2017inverse}
D.~Hadfield-Menell, S.~Milli, P.~Abbeel, S.~J. Russell, and A.~Dragan,
  ``Inverse reward design,'' \emph{Advances in neural information processing
  systems}, vol.~30, 2017.

\bibitem{fu2018learning}
\BIBentryALTinterwordspacing
J.~Fu, K.~Luo, and S.~Levine, ``Learning robust rewards with adverserial
  inverse reinforcement learning,'' in \emph{International Conference on
  Learning Representations}, 2018. [Online]. Available:
  \url{https://openreview.net/forum?id=rkHywl-A-}
\BIBentrySTDinterwordspacing

\bibitem{everitt2021reward}
T.~Everitt, M.~Hutter, R.~Kumar, and V.~Krakovna, ``Reward tampering problems
  and solutions in reinforcement learning: A causal influence diagram
  perspective,'' \emph{Synthese}, vol. 198, no.~27, pp. 6435--6467, 2021.

\bibitem{rajeswaran2017learning}
A.~Rajeswaran, V.~Kumar, A.~Gupta, G.~Vezzani, J.~Schulman, E.~Todorov, and
  S.~Levine, ``Learning complex dexterous manipulation with deep reinforcement
  learning and demonstrations,'' \emph{arXiv preprint arXiv:1709.10087}, 2017.

\bibitem{zhu2018reinforcement}
Y.~Zhu, Z.~Wang, J.~Merel, A.~Rusu, T.~Erez, S.~Cabi, S.~Tunyasuvunakool,
  J.~Kram{\'a}r, R.~Hadsell, N.~de~Freitas \emph{et~al.}, ``Reinforcement and
  imitation learning for diverse visuomotor skills,'' \emph{arXiv preprint
  arXiv:1802.09564}, 2018.

\bibitem{farmer2009economy}
J.~D. Farmer and D.~Foley, ``The economy needs agent-based modelling,''
  \emph{Nature}, vol. 460, no. 7256, pp. 685--686, 2009.

\bibitem{suo2021trafficsim}
S.~Suo, S.~Regalado, S.~Casas, and R.~Urtasun, ``Trafficsim: Learning to
  simulate realistic multi-agent behaviors,'' in \emph{ICCV}, 2021.

\bibitem{igl2022symphony}
M.~Igl, D.~Kim, A.~Kuefler, P.~Mougin, P.~Shah, K.~Shiarlis, D.~Anguelov,
  M.~Palatucci, B.~White, and S.~Whiteson, ``Symphony: Learning realistic and
  diverse agents for autonomous driving simulation,'' in \emph{ICRA}, 2022.

\bibitem{grover2018learning}
A.~Grover, M.~Al-Shedivat, J.~Gupta, Y.~Burda, and H.~Edwards, ``Learning
  policy representations in multiagent systems,'' in \emph{International
  conference on machine learning}.\hskip 1em plus 0.5em minus 0.4em\relax PMLR,
  2018, pp. 1802--1811.

\bibitem{liang2020agent}
Y.~Liang, C.~Guo, Z.~Ding, and H.~Hua, ``Agent-based modeling in electricity
  market using deep deterministic policy gradient algorithm,'' \emph{IEEE
  Transactions on Power Systems}, vol.~35, no.~6, 2020.

\bibitem{wang2017robust}
Z.~Wang, J.~S. Merel, S.~E. Reed, N.~de~Freitas, G.~Wayne, and N.~Heess,
  ``Robust imitation of diverse behaviors,'' \emph{NeurIPS}, 2017.

\bibitem{lucic2018gans}
M.~Lucic, K.~Kurach, M.~Michalski, S.~Gelly, and O.~Bousquet, ``Are gans
  created equal? a large-scale study,'' \emph{NeurIPS}, 2018.

\bibitem{creswell2018generative}
A.~Creswell, T.~White, V.~Dumoulin, K.~Arulkumaran, B.~Sengupta, and A.~A.
  Bharath, ``Generative adversarial networks: An overview,'' \emph{IEEE signal
  processing magazine}, vol.~35, no.~1, pp. 53--65, 2018.

\bibitem{goodfellow2020generative}
I.~Goodfellow, J.~Pouget-Abadie, M.~Mirza, B.~Xu, D.~Warde-Farley, S.~Ozair,
  A.~Courville, and Y.~Bengio, ``Generative adversarial networks,''
  \emph{Communications of the ACM}, vol.~63, no.~11, pp. 139--144, 2020.

\bibitem{ho2016generative}
J.~Ho and S.~Ermon, ``Generative adversarial imitation learning,''
  \emph{NeurIPS}, 2016.

\bibitem{baram2016model}
N.~Baram, O.~Anschel, and S.~Mannor, ``Model-based adversarial imitation
  learning,'' \emph{arXiv preprint arXiv:1612.02179}, 2016.

\bibitem{xu2022bits}
D.~Xu, Y.~Chen, B.~Ivanovic, and M.~Pavone, ``Bits: Bi-level imitation for
  traffic simulation,'' \emph{arXiv preprint arXiv:2208.12403}, 2022.

\bibitem{ross2011reduction}
S.~Ross, G.~Gordon, and D.~Bagnell, ``A reduction of imitation learning and
  structured prediction to no-regret online learning.''\hskip 1em plus 0.5em
  minus 0.4em\relax JMLR Workshop and Conference Proceedings, 2011.

\bibitem{ettinger2021large}
S.~Ettinger, S.~Cheng, B.~Caine, C.~Liu, H.~Zhao, S.~Pradhan, Y.~Chai, B.~Sapp,
  C.~Qi, Y.~Zhou, Z.~Yang, A.~Chouard, P.~Sun, J.~Ngiam, V.~Vasudevan,
  A.~McCauley, J.~Shlens, and D.~Anguelov, ``Large scale interactive motion
  forecasting for autonomous driving : The waymo open motion dataset,''
  \emph{CoRR}, 2021.

\bibitem{kingma2013auto}
D.~P. Kingma and M.~Welling, ``Auto-encoding variational bayes,'' \emph{ICLR},
  2014.

\bibitem{lynch2020learning}
C.~Lynch, M.~Khansari, T.~Xiao, V.~Kumar, J.~Tompson, S.~Levine, and
  P.~Sermanet, ``Learning latent plans from play,'' in \emph{Conference on
  robot learning}.\hskip 1em plus 0.5em minus 0.4em\relax PMLR, 2020, pp.
  1113--1132.

\bibitem{de2019causal}
P.~De~Haan, D.~Jayaraman, and S.~Levine, ``Causal confusion in imitation
  learning,'' \emph{NeurIPS}, 2019.

\bibitem{schulman2015trust}
J.~Schulman, S.~Levine, P.~Abbeel, M.~Jordan, and P.~Moritz, ``Trust region
  policy optimization,'' in \emph{ICML}, 2015.

\bibitem{schulman2017proximal}
J.~Schulman, F.~Wolski, P.~Dhariwal, A.~Radford, and O.~Klimov, ``Proximal
  policy optimization algorithms,'' \emph{arXiv:1707.06347}, 2017.

\bibitem{sutton1999between}
R.~S. Sutton, D.~Precup, and S.~Singh, ``Between mdps and semi-mdps: A
  framework for temporal abstraction in reinforcement learning,''
  \emph{Artificial intelligence}, vol. 112, no. 1-2, pp. 181--211, 1999.

\bibitem{bacon2017option}
P.-L. Bacon, J.~Harb, and D.~Precup, ``The option-critic architecture,'' in
  \emph{AAAI}, 2017.

\bibitem{vezhnevets2017feudal}
A.~S. Vezhnevets, S.~Osindero, T.~Schaul, N.~Heess, M.~Jaderberg, D.~Silver,
  and K.~Kavukcuoglu, ``Feudal networks for hierarchical reinforcement
  learning,'' in \emph{ICML}, 2017.

\bibitem{nachum2018nearoptimal}
O.~Nachum, S.~Gu, H.~Lee, and S.~Levine, ``Near-optimal representation learning
  for hierarchical reinforcement learning,'' in \emph{ICLR}, 2019.

\bibitem{igl2020multitask}
M.~Igl, A.~Gambardella, J.~He, N.~Nardelli, N.~Siddharth, W.~B{\"o}hmer, and
  S.~Whiteson, ``Multitask soft option learning,'' in \emph{UCA}, 2020.

\bibitem{krishnan2017ddco}
S.~Krishnan, R.~Fox, I.~Stoica, and K.~Goldberg, ``Ddco: Discovery of deep
  continuous options for robot learning from demonstrations,'' in
  \emph{Conference on robot learning}.\hskip 1em plus 0.5em minus 0.4em\relax
  PMLR, 2017, pp. 418--437.

\bibitem{le2018hierarchical}
H.~Le, N.~Jiang, A.~Agarwal, M.~Dudik, Y.~Yue, and H.~Daum{\'e}~III,
  ``Hierarchical imitation and reinforcement learning,'' in \emph{ICML}, 2018.

\bibitem{shiarlis2018taco}
K.~Shiarlis, M.~Wulfmeier, S.~Salter, S.~Whiteson, and I.~Posner, ``Taco:
  Learning task decomposition via temporal alignment for control,'' in
  \emph{ICML}, 2018.

\bibitem{merel2017learning}
J.~Merel, Y.~Tassa, D.~TB, S.~Srinivasan, J.~Lemmon, Z.~Wang, G.~Wayne, and
  N.~Heess, ``Learning human behaviors from motion capture by adversarial
  imitation,'' \emph{arXiv:1707.02201}, 2017.

\bibitem{fei2020triple}
C.~Fei, B.~Wang, Y.~Zhuang, Z.~Zhang, J.~Hao, H.~Zhang, X.~Ji, and W.~Liu,
  ``Triple-gail: a multi-modal imitation learning framework with generative
  adversarial nets,'' \emph{arXiv preprint arXiv:2005.10622}, 2020.

\bibitem{tamar2018imitation}
A.~Tamar, K.~Rohanimanesh, Y.~Chow, C.~Vigorito, B.~Goodrich, M.~Kahane, and
  D.~Pridmore, ``Imitation learning from visual data with multiple
  intentions,'' in \emph{ICLR}, 2018.

\bibitem{khandelwal2020if}
S.~Khandelwal, W.~Qi, J.~Singh, A.~Hartnett, and D.~Ramanan, ``What-if motion
  prediction for autonomous driving,'' \emph{arXiv preprint arXiv:2008.10587},
  2020.

\bibitem{ding2019goal}
Y.~Ding, C.~Florensa, P.~Abbeel, and M.~Phielipp, ``Goal-conditioned imitation
  learning,'' \emph{NeurIPS}, 2019.

\bibitem{pashevich2021episodic}
A.~Pashevich, C.~Schmid, and C.~Sun, ``Episodic transformer for
  vision-and-language navigation,'' in \emph{ICCV}, 2021.

\bibitem{vinyals2019grandmaster}
O.~Vinyals, I.~Babuschkin, W.~M. Czarnecki, M.~Mathieu, A.~Dudzik, J.~Chung,
  D.~H. Choi, R.~Powell, T.~Ewalds, P.~Georgiev \emph{et~al.}, ``Grandmaster
  level in starcraft ii using multi-agent reinforcement learning,''
  \emph{Nature}, vol. 575, no. 7782, pp. 350--354, 2019.

\bibitem{li2017infogail}
Y.~Li, J.~Song, and S.~Ermon, ``Infogail: Interpretable imitation learning from
  visual demonstrations,'' \emph{Advances in Neural Information Processing
  Systems}, vol.~30, 2017.

\bibitem{NIPS2017_632cee94}
K.~Hausman, Y.~Chebotar, S.~Schaal, G.~Sukhatme, and J.~J. Lim, ``Multi-modal
  imitation learning from unstructured demonstrations using generative
  adversarial nets,'' in \emph{NeurIPS}, 2017.

\bibitem{eysenbach2018diversity}
B.~Eysenbach, A.~Gupta, J.~Ibarz, and S.~Levine, ``Diversity is all you need:
  Learning skills without a reward function,'' in \emph{ICLR}, 2019.

\bibitem{pathak2019self}
D.~Pathak, D.~Gandhi, and A.~Gupta, ``Self-supervised exploration via
  disagreement,'' in \emph{ICML}, 2019.

\end{thebibliography}
\bibliographystyle{IEEEtran}

\end{document}